# Enhancing Vaccine Safety Surveillance: Extracting Vaccine Mentions from Emergency Department Triage Notes Using Fine-Tuned Large Language Models

Sedigh KHADEMI[a,1], Jim BLACK[b], Christopher PALMER[a], Muhammad JAVED[a], Hazel Clothier[a], Jim BUTTERY[a,c], and Gerardo Luis DIMAGUILA[a]

[a] *Murdoch Children's Research Institute, Parkville, Australia*
[b] *Department of Health, State Government of Victoria, Melbourne, Australia*
[c] *Department of Paediatrics, The University of Melbourne, Parkville, Australia*

ORCiD ID: Sedigh Khademi https://orcid.org/0000-0001-6146-1415 , Jim Black https://orcid.org/0000-0002-9287-8712 , Christopher Palmer https://orcid.org/0000-0001-6554-9027, Muhammad Javed https://orcid.org/0000-0002-7022-6596, Hazel Clothier https://orcid.org/0000-0001-7594-0361,  Jim Buttery https://orcid.org/0000-0001-9905-2035,
Gerardo Luis Dimaguila https://orcid.org/0000-0002-3498-6256

**Abstract.** This study evaluates fine-tuned Llama 3.2 models for extracting vaccine-related information from emergency department triage notes to support near real-time vaccine safety surveillance. Prompt engineering was used to initially create a labeled dataset, which was then confirmed by human annotators. The performance of prompt-engineered models, fine-tuned models, and a rule-based approach was compared. The fine-tuned Llama 3 billion parameter model outperformed other models in its accuracy of extracting vaccine names. Model quantization enabled efficient deployment in resource-constrained environments. Findings demonstrate the potential of large language models in automating data extraction from emergency department notes, supporting efficient vaccine safety surveillance and early detection of emerging adverse events following immunization issues.

## 1. Introduction

Extracting vaccine-related health information from electronic health records (EHRs), especially emergency department (ED) triage notes, provides a valuable resource for enabling efficient, near real-time surveillance of vaccine safety, complementing traditional passive surveillance methods. Adverse Events Following Immunization (AEFI) encompass any untoward medical occurrences that follow immunization, regardless of whether a causal relationship with the vaccine exists. Natural Language

---

[1]Corresponding Author: Sedigh Khademi, sedigh.khademi@mcri.edu.au, Murdoch Children's Research Institute, 50 Flemington Road, Parkville VIC 3052, Australia

Processing (NLP) provides effective methods for extracting clinical information from unstructured clinical notes [1]. Developing these models typically demands annotated datasets and considerable effort to align extracted information with predefined classifications [2].

The advent of large language models (LLMs) has introduced new opportunities for automating data annotation and enhancing information extraction tasks[3,4]. Instruction-tuning these models allows for better alignment of extracted text with expected outputs, streamlining the process and reducing manual intervention[5]. However, deployment of LLMs in real-world applications necessitates consideration of computational efficiency and adaptability, especially in resource-constrained environments.

In this study, we developed a fine-tuned Llama 3.2 model for an AEFI pipeline monitoring ED triage notes. The LLM extracted vaccine names from potential AEFI-containing notes. Using prompt engineering, we generated an annotated dataset, validated by human annotators, and used to fine-tune the Llama models. We compared prompt engineered vs fine-tuned models, and a rule-based approach.

## 2. Methods

*2.1. Dataset*

SynSurv, the syndromic surveillance project of Victoria's Department of Health in Australia, serves as the primary data source for this study. SynSurv collects near-real-time emergency department (ED) presentation data from most public hospitals in Victoria, Australia. Most data arrive to the SynSurv system within 5–15 minutes of patient assessment. The triage text, created by ED nurses, consists of abbreviated phrases describing the patient's presenting complaint, medical history, and observations, without including personal identifying details. The text length can vary from a detailed narrative to a brief diagnostic summary. Age and gender fields are also included in the data. ED triage text is very idiosyncratic compared to the more formal text found in typical clinical records – we have found that existing NLP applications do not work well with this text and models trained specifically for the text are required.

Using a Transformer-based NLP model [6], SynSurv identifies ED triage notes that mention potential adverse events following a recent vaccination. For this study, we used 589 (61 non-vaccine, 528 vaccine-related) of these notes for model training and an additional 259 notes (238 non-vaccine, 21 vaccine-related) for testing. The training dataset was initially submitted to the Llama-3B (three billion parameters) model using prompt engineering, which provided a benchmark for the model's performance, and assisted with labeling - which was completed by two authors. Table 1 provides examples of triage notes and the vaccination name we aim to extract.

**Table 1.** Sample of triage notes (Illustrative examples)

| Triage Note | Vaccine |
|---|---|
| Age: 0Y 1M. 6wo vaccinations yesterday, eye swelling, redness and wob post. Afebrile. | 6 weeks |
| Age: 13Y 2 M. Allergic reaction post imms. Rash to neck, felt throat closing over. | Unspecified |
| Age: 0Y 4M. Febrile, blood in stool, vomit post rota-virus vaccine | Rotavirus |
| Age: 3Y 1M. fever, runny nose, sob on b/g of flu vax 2/7 ago | Influenza |
| Age: 5M. whooping cough prophylaxis 2/52 ago, 4/7 of fevers, increased sob and coughs | No |

*2.2. Models*

We used Llama 3.2 1B and 3B models to compare prompt engineering and fine-tuned approaches, evaluating their performance against a baseline rule-based algorithm. For fine-tuning we used the Hugging Face PEFT (Parameter-Efficient-Fine-Tuning) library with QLoRA (Quantized Low-Rank Adaptation), which works by fine-tuning over a smaller number of the trainable parameters and reducing model resource requirements via quantization. This preserves the model's innate knowledge and allows efficient and cost-effective fine-tuned and deployable models using single consumer-grade GPUs, which is not possible with the original model. Fine tuning acquaints the LLM with both the target domain and with the expected responses as labels, so a fine-tuned model is likely to require less post-processing to normalize its responses. Figure 1 shows the prompt used for the initial prompt-engineered 3B model, and which was adapted for the prompting technique used for fine-tuning.

```
You are an expert medical analyst reviewing emergency department triage notes.
Your role is to analyze the provided note with precision, answering the following questions concisely and
accurately based on the given triage note.
The triage note starts with patient age and a chief complaint.
Use expert medical reasoning and comprehensive categorization to complete the task.
Provide responses in a structured JSON format.
Include the analysis as follows:
Vaccination:
  Extract the name or type of vaccine mentioned in the triage notes.
  If a specific vaccine is named (e.g., "Influenza", "COVID-19"), return only the name.
  Only use vaccine names explicitly mentioned in the text; do not make up or infer vaccine names.
  If an injection, immunization, or shot of a vaccine is mentioned without specifying a vaccine name,
    respond with "Unspecified" if it implies a general vaccination.
  If the injection or shot is not explicitly a vaccine (e.g., other medications or treatments),
    respond with "No"
  If the note mentions that the person is scheduled or due for an immunization but has not yet received
    it, respond with "No".
  If there is no mention of any vaccine, vaccination, or injection, respond with "No".
  Only provide the exact vaccine name if clearly specified in the notes.
  Do not infer or assume the vaccine name based on symptoms or other text details.
  Responses should be limited to the exact vaccine name or the designated response (Unspecified or No).
```

**Figure 1.** Prompt used for extracting vaccine.

## 3. Results

To quantify the model's success in standard terms of precision, recall and F1-score, we simplified the models' predictions to whether they successfully identified the presence or not of a vaccine. Scores are shown in table 2.

**Table 2**: Models' standard scoring

| Model | TP | TN | FN | FP | Precision | Recall | F1 |
|---|---|---|---|---|---|---|---|
| Llama 1B prompt engineering | 212 | 14 | 26 | 7 | 0.97 | 0.89 | 0.93 |
| Llama 1B Fine-tuned | 232 | 14 | 6 | 7 | 0.97 | 0.98 | 0.97 |
| Llama 3B Fine-tuned | 225 | 18 | 13 | 3 | 0.99 | 0.95 | 0.97 |
| Rules-based | 219 | 18 | 19 | 6 | 0.99 | 0.93 | 0.95 |

The fine-tuned models significantly reduced false negatives compared to the prompted model, achieving higher recall and precision with more true positives. While the fine-tuned 1B model produced more false positives than the 3B model, its lower rate of false negatives makes it a potentially better choice for naming vaccines where AEFI are potentially associated. The fine-tuned Llama models demonstrated a modest performance improvement over a baseline rules-based approach. However, the rules-based method required significantly more development effort and ongoing maintenance, making it less favorable despite its high specificity.

We evaluated the models' accuracy in extracting vaccine names separately, given variability in both the data and the LLMs' identification of vaccine names. For instance, terms like 'Hep B,' 'Hepatitis B,' and various brand names were all valid identifiers for the hepatitis B vaccine. Similarly, while 'Influenza' was the preferred term for the influenza vaccine, alternatives such as 'flu vax' or 'flu shot' were acceptable, as they reflected the language found in the text. We also accounted for the LLMs' inherent knowledge; for example, a response of 'Triple Antigen' for a mention of 'DTP,' both of which are valid identifiers for the diphtheria, tetanus, and whooping cough vaccine.

Table 3 shows the models' predictions for Unspecified vaccine and all vaccines. 'Correct' indicates accurate vaccine identification, while 'Incorrect' represents errors. The fine-tuned 1B model performed best for 'Unspecified' labels, while the fine-tuned 3B model was the best overall performer.

**Table 3**: Examples of models' accuracy in identifying vaccines

| Label | Unspecified (107) | | | All (259) | | |
|---|---|---|---|---|---|---|
| Response | Correct | Incorrect | % Correct | Correct | Incorrect | % Correct |
| 1B prompted | 88 | 19 | 82% | 178 | 81 | 68.7% |
| 1B fine-tuned | 100 | 7 | 93% | 211 | 48 | 81.5% |
| 3B fine-tuned | 92 | 15 | 86% | 224 | 35 | 86.5% |

When the models' responses exactly matched the supplied label, less post-processing was required to normalize responses. The 3B model achieved 86.5% (224/259) exact matching compared to the 1B model's 79.9% (207/259).

## 4. Discussion

In this study, we developed a fine-tuned Llama-based model as part of a pipeline for extracting vaccine names from vaccine-related triage notes. By leveraging prompt engineering, we created an initial annotated dataset, which was validated by human annotators and subsequently used to fine-tune LLMs. These models outperformed the baseline rule-based approach, and with more potential to adapt to changing vaccine-related ED data. Furthermore, the use of fine-tuned LLMs is more suitable than a rules-based approach for other named entity recognition tasks such as identifying symptoms. We observed that quantization, which enhances the models' suitability for resource-constrained settings, did not affect the models' vaccine identification performance.

Our study provides valuable insights into leveraging LLMs for information extraction from medical records. We demonstrated that combining prompt engineering techniques with human annotators can substantially reduce the effort required for manual annotation, facilitating the efficient and rapid development of labeled datasets. This approach aligns with and is validated by findings from other studies, which have similarly highlighted the effectiveness of integrating LLMs and human expertise to streamline data annotation processes [7,8]. Additionally, we show that small Llama 3.2

LLMs, fine-tuned on a compact dataset, are capable of both extraction and normalization tasks. These models are deployable on CPU-based machines through quantization, which maintains model performance while making them suitable for more resource-constrained environments. Our study has limitations. First, we did not evaluate other small LLMs. Additionally, since the LLMs were not trained on clinical text data, further training on such data could potentially enhance the model's performance. While our training and test datasets show an imbalance in vaccine-related and non-vaccine-related notes, potentially risking overfitting especially for individual vaccine types, the operationalized model has maintained consistent performance in real-world deployment where it encounters more varied data. As we enhance the model, we will expand our dataset to be more balanced. Moreover, our limited training and evaluation data restricted the scope of our analysis. Broader data diversity, along with the use of active learning techniques to build high-quality datasets [9], could lead to more generalizable results.

## 5. Conclusions

Syndromic surveillance of emergency department notes for potential vaccine adverse events and the vaccines involved, is an important contribution to early detection of emerging issues with a vaccination schedule. Natural Language Processing assists with automation of AEFI detection and recognition of the vaccines involved, but requires models specifically adapted to the types of text found in ED notes. In this paper we have evaluated the use of fine-tuned LLMs and have found them to be very effective at identifying the vaccines involved in an AEFI.